\documentclass[a4paper,conference]{IEEEtran}
\ifCLASSINFOpdf
\else
\fi
\usepackage{graphicx, subfigure, amsmath}
\usepackage[T1]{fontenc}
\usepackage{fancynum}

\begin{document}
%
% paper title
% Titles are generally capitalized except for words such as a, an, and, as,
% at, but, by, for, in, nor, of, on, or, the, to and up, which are usually
% not capitalized unless they are the first or last word of the title.
% Linebreaks \\ can be used within to get better formatting as desired.
% Do not put math or special symbols in the title.
\title{Detecting Heads using Feature Refine Net and Cascaded Multi-scale Architecture}

% author names and affiliations
% use a multiple column layout for up to three different
% affiliations
\author{
	{Dezhi Peng${}^{*}$, Zikai Sun${}^{*}$, Zirong Chen, Zirui Cai, Lele Xie, Lianwen Jin}\\
	{School of Electronic and Information Engineering}\\
	{South China University of Technology}\\
	{\{eedzpeng, eezksun, eezrchen\}}@mail.scut.edu.cn and {\{cai.zirui, xie.lele\}}@foxmail.com and eelwjin@scut.edu.cn
}

% conference papers do not typically use \thanks and this command
% is locked out in conference mode. If really needed, such as for
% the acknowledgment of grants, issue a \IEEEoverridecommandlockouts
% after \documentclass

% for over three affiliations, or if they all won't fit within the width
% of the page, use this alternative format:
%
%\author{\IEEEauthorblockN{Michael Shell\IEEEauthorrefmark{1},
%Homer Simpson\IEEEauthorrefmark{2},
%James Kirk\IEEEauthorrefmark{3},
%Montgomery Scott\IEEEauthorrefmark{3} and
%Eldon Tyrell\IEEEauthorrefmark{4}}
%\IEEEauthorblockA{\IEEEauthorrefmark{1}School of Electrical and Computer Engineering\\
%Georgia Institute of Technology,
%Atlanta, Georgia 30332--0250\\ Email: see http://www.michaelshell.org/contact.html}
%\IEEEauthorblockA{\IEEEauthorrefmark{2}Twentieth Century Fox, Springfield, USA\\
%Email: homer@thesimpsons.com}
%\IEEEauthorblockA{\IEEEauthorrefmark{3}Starfleet Academy, San Francisco, California 96678-2391\\
%Telephone: (800) 555--1212, Fax: (888) 555--1212}
%\IEEEauthorblockA{\IEEEauthorrefmark{4}Tyrell Inc., 123 Replicant Street, Los Angeles, California 90210--4321}}

% use for special paper notices
%\IEEEspecialpapernotice{(Invited Paper)}

% make the title area
\maketitle

\renewcommand{\thefootnote}{}

\footnotetext[1]{
\rule[0.01cm]{1.5cm}{0.01cm}

${}^{*}$Dezhi Peng and Zikai Sun make equal contribution to this work.}	

\newcommand{\tabincell}[2]{\begin{tabular}{@{}#1@{}}#2\end{tabular}}  
% As a general rule, do not put math, special symbols or citations
% in the abstract
\begin{abstract}
This paper presents a method that can accurately detect heads especially small heads under the indoor scene. To achieve this, we propose a novel method, Feature Refine Net (FRN), and a cascaded multi-scale architecture. FRN exploits the multi-scale hierarchical features created by deep convolutional neural networks. The proposed channel weighting method enables FRN to make use of features alternatively and effectively. To improve the performance of small head detection, we propose a cascaded multi-scale architecture which has two detectors. One called global detector is responsible for detecting large objects and acquiring the global distribution information. The other called local detector is designed for small objects detection and makes use of the information provided by global detector. Due to the lack of head detection datasets, we have collected and labeled a new large dataset named SCUT-HEAD which includes 4405 images with 111251 heads annotated. Experiments show that our method has achieved state-of-the-art performance on SCUT-HEAD.
\end{abstract}
	
% no keywords

% For peer review papers, you can put extra information on the cover
% page as needed:
% \ifCLASSOPTIONpeerreview
% \begin{center} \bfseries EDICS Category: 3-BBND \end{center}
% \fi
%
% For peerreview papers, this IEEEtran command inserts a page break and
% creates the second title. It will be ignored for other modes.
\maketitle

\section{Introduction}
% no \IEEEPARstart
Face detection and pedestrian detection are two important research problems in computer vision and significant results have been achieved in recent years. However, there are some limitations in practical application. Face detection can only detect faces, which means a person who turns his back to the camera can not be detected. Due to the complexity of the indoor scene, most parts of body are not visible. Therefore, pedestrian detection is also hard to work in such situation. Head detection doesn't have these limitations, hence is more suitable for people locating and counting, especially under the indoor scene. However, there are also many challenges to detect heads under the indoor scene such as the variance of scales and appearances of heads, and small head detection.

Due to the various scales and appearances of heads, how to exploit extracted features effectively to localize heads and distinguish them from background remains a big problem. Many previous methods make use of multi-scale features generated at different levels of deep convolutional neural networks. Hariharan \textsl{et al.}\cite{hypercolumn} encodes concatenated rescaled feature maps at different levels into one vector called hypercolumn for every location. SSD\cite{ssd} makes effort to employ multi-scale features to estimate class probability and bounding box coordinates. Lin \textsl{et al.}\cite{fpn} proposes a top-down architecture for building high-level semantics feature maps of different scales and make predictions on feature maps of different scales respectively. Some other methods such as HyperNet\cite{hypernet} and ParseNet\cite{parsenet} combine multiple layers together for the final predictions. Many experiments have implied that making use of multi-scale features makes sense. In this paper, we propose a novel method named Feature Refine Net (FRN) for exploiting multi-scale features. Compared to previous methods, FRN uses channel weighting to perform feature selection by adding learnable weights to channels of feature maps. The most useful features for the specific domain are selected and made use of. Moreover, feature decomposition upsampling is proposed to upsample small feature maps by decomposing one pixel to a related region. Resized feature maps are concatenated and undergo an Inception-style synthesis. Experiments have proven that FRN provides a great improvement on detection performance.

Small heads detection is another problem that must be addressed. Hu \textsl{et al.}\cite{HR} proposes a framework named HR which resizes the input image to different scales and applies scale invariant detectors. Inspired by attention mechanism of human, we proposed a cascaded multi-scale architecture for small heads detection. Rather than resizing the entire image to different scales like HR, our method focuses on refining local detection results by increasing the resolution of clips of an image. The proposed architecture consists of two different detectors named global detector and local detector respectively. Global detector detects large heads and informs local detector about the location of small heads. Then local detector works on the enlarged clips which contain small heads for more accurate small head detection.

Due to the lack of head detection datasets, we have also collected and labeled a large-scale head detection dataset named SCUT-HEAD. Our method reaches 0.91 Hmean on partA and 0.90 Hmean on partB, which outperforms many popular object detection frameworks such as Faster R-CNN\cite{faster-rcnn}, R-FCN\cite{rfcn}, YOLO\cite{yolo9000} and SSD\cite{ssd}.
	
\begin{figure*}[!htb]
	\centering
	\includegraphics[width=1.0\textwidth]{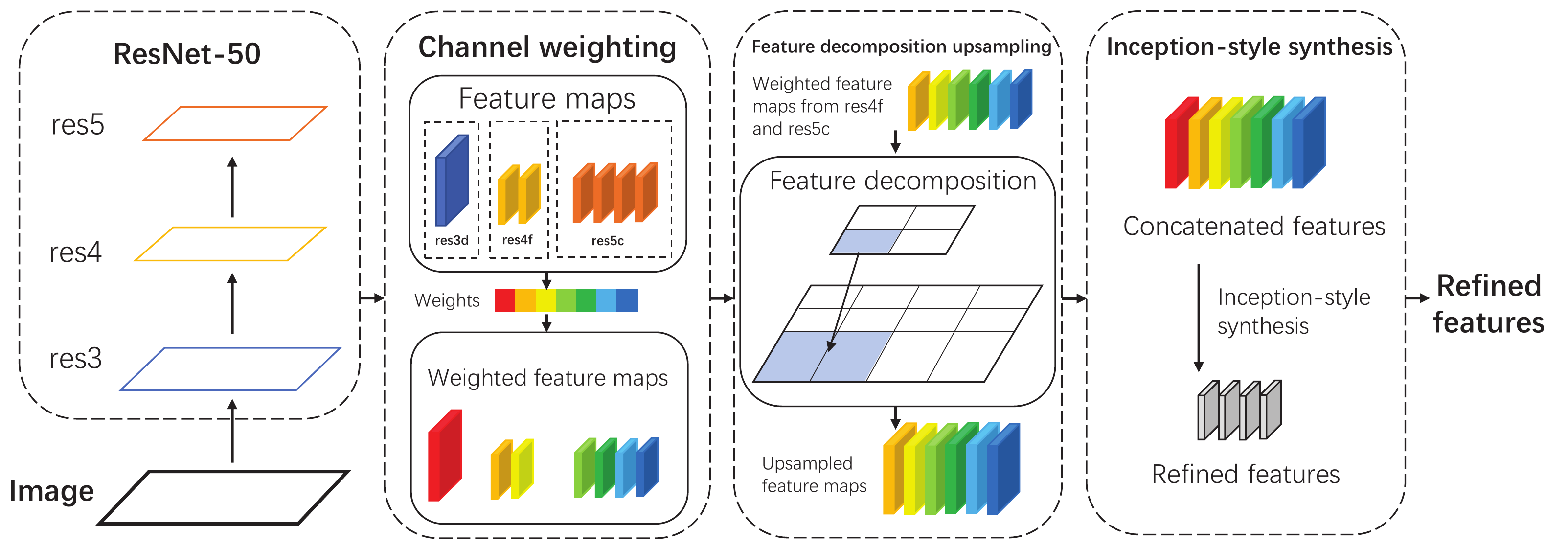}
	\caption{The overall architecture of FRN (based on ResNet-50): (1) Channel weighting is applied on res3, res4 and res5 to perform feature selection. (2) Weighted features undergo feature decomposition upsampling and their scales are increased twofold. (3) Three groups of feature maps are concatenated together along the channel dimension. (4) We adopt Inception-style synthesis method to composite the concatenated feature maps in order to make use of the internal relationship between channels and reduce the computation complexity.}\label{frn}
\end{figure*}
To summarize, the main contributions of this paper are listed as follows:
\begin{itemize}
	\setlength{\itemsep}{0pt}
	\setlength{\parsep}{0pt}
	\setlength{\parskip}{0pt}
	\item We propose a new model named Feature Refine Net (FRN) for multi-scale features combination and automatic feature selection.
	\item A cascaded multi-scale architecture is designed for small heads detection.
	\item A head detection dataset named SCUT-HEAD with 4405 images and 111251 annotated heads is built.
\end{itemize}

\section{Method}
\label{sec:method}
	
\subsection{Overall Architecture}
In this paper, we implement our method based on R-FCN\cite{rfcn} and use ResNet-50 (ignoring pool5, fc1000 and prob layers) as feature extractor. We denote the feature maps produced by res3x, res4x and res5x blocks as res3, res4 and res5 respectively. FRN shown in Fig.\ref{frn} is inserted into R-FCN framework and RPN works on the output of FRN for region proposals. Then we train two modified R-FCNs named local detector and global detector for the cascaded multi-scale architecture which is shown in Fig.\ref{cma}. The cascaded multi-scale architecture consists of four stages of (1) a global detector that works on the entire image to detect large heads and obtains the rough location of small heads; (2) multiple clips which have high probability of containing small heads; (3) a local detector that works on the clips and results in more accurate head detection; (4) an ensemble module that merges both local and global detectors and non maximum suppression is applied.
	
\subsection{Feature Refine Net}
\label{sec:fmn}
Feature Refine Net (FRN) refines the multiple feature maps res3, res4 and res5. Firstly, through channel weighting, each channel of feature maps is multiplied by the corresponding learnable weight. Then, we use feature decomposition upsampling to increase the resolution of res4 and res5 twofold. Next, feature maps are concatenated along the channel dimension. Finally, concatenated feature maps undergo Inception-style synthesis yielding refined features.
	
\subsubsection{Channel Weighting}
%\subsubsection{Channel Weighting}
Deep convolutional neural networks generate multiple feature maps at different layers. The feature maps generated at low levels contain more detailed information and have a smaller receptive field, hence are more suitable for small object detection and precise locating. The feature maps generated at high levels contain more abstract but coarser information and have a larger receptive field. Therefore they are suitable for large object detection and classification. Due to the different characteristics mentioned above, the selection of feature maps will be useful. The feature extractor pre-trained on ImageNet\cite{imagenet} such as VGG\cite{vgg}, ResNet\cite{resnet} has proven to have great generalizing ability, yielding general representation of objects. However, even after finetuning, extracted features still reserve some characteristics of object categories in ImageNet. Direct usage of features may not be the best choice. Thus we use channel weighting to select and take advantage of the most useful features.

Channel weighting is the key component of FRN. We multiply each channel of feature maps with the corresponding learnable weight parameter. It enables FRN to select which feature to use automatically, which means the detector with FRN will be more adaptive for the specific domain. Let $i$ denote the index of channel, $j,k$ denote the spatial position of pixels in a feature map and $N$ denote the number of channels. The relationship between input feature maps $f_{C \times W \times H}$ and output feature maps $F_{C \times W \times H}$ can be expressed as follows:
	
\begin{equation}
F_{i,j,k} = w_i \cdot f_{i,j,k} , i = 1, 2, ..., N 
\end{equation}

\noindent where $w_i$ is the weight parameter. Weight parameters are optimized during backpropagating. Let $Loss^F$ denote the loss of the output feature maps $F$. Then the gradients of $Loss_F$ with respect to $w_i$ and $f_{i,j,k}$ are as follows:

\begin{equation}
\frac{\partial Loss^F}{\partial w_i} = \sum_{j = 1}^{W} \sum_{k = 1}^{H} f_{i,j,k}
\end{equation}

\begin{equation}
\frac{\partial Loss^F}{\partial f_{i,j,k}} = w_i
\end{equation}
	
In our method, we apply channel weighting to res3, res4 and res5 respectively. As shown in Fig.\ref{wc}, the channels which contain more useful information have higher weights. From the analysis in section \ref{results} and \ref{channel weighting}, channel weighting performs feature selection very well and raises the accuracy as well.
	
\subsubsection{Feature Decomposition Upsampling} 
%\subsubsection{Position Sensitive upsampling}
Previous methods such as \cite{fpn,parsenet} use nearest neighbor upsampling or bilinear interpolation or even simply replication to upsample small feature maps. Unlike previous methods, we conduct feature maps upsampling by feature decomposition. Every pixel in a feature map is related to a local region of feature maps at low level. Therefore, we decompose each pixel to a $N \times N$ region to upsample a feature map. We use a mapping matrix $M_{N \times N}$ to represent the relationship between the input pixel $p$ and the decomposed  $N \times N$ region $P_{N \times N}$.

\begin{equation}
P_{N \times N} = p \cdot M_{N \times N}
\end{equation} 

Because each channel of feature maps represent a specific feature of an object, we use different mapping matrices for different channels. The relationship between the upsampled feature maps $F_{C \times WN \times HN}$ and the input feature maps $f_{C \times W \times H}$ can be expressed as follows:

\begin{equation}
F_{i,j,k} = M_{i,(j\bmod N),(k\bmod N)} \cdot f_{i, \lfloor \frac{j}{N} \rfloor, \lfloor \frac{k}{N} \rfloor}
\end{equation}  

Let $Loss^F$ denote the loss of the output feature maps $F$. Then during backpropagation, the gradients with respect to $M_{i,m,n}$ and $f_{i,x,y}$ are as follows:

\begin{equation}
\frac{\partial Loss^F}{\partial M_{i,m,n}} = \sum_{x = 1}^{W} \sum_{y = 1}^{H} f_{i,x,y}
\end{equation}

\begin{equation}
\frac{\partial Loss^F_{i,j,k}}{\partial f_{i,x,y}} = \begin{cases}
M_{i,(j\bmod N),(k\bmod N)}, & \lfloor \frac{j}{N} \rfloor = x, \lfloor \frac{k}{N} \rfloor = y\\
0, & otherwise
\end{cases}
\end{equation}

In our method, the weighted feature maps from res4, res5 are upsampled to match the scale of res3. Every pixel is decomposed to a $2 \times2$ region according to the corresponding mapping matrix.
	
\subsubsection{Inception-Style Synthesis}
%\subsubsection{Inception-Style Synthesis}
After upsampling, multiple feature maps are concatenated together along the channel dimension. However, simple concatenating operation has two drawbacks. Firstly, concatenated feature maps have too many channels and their scale is too large. It is a big challenge for computation capacities if we simply apply concatenated feature maps to the following process. Secondly, concatenated feature maps lack the internal relationship between different channels because all the previous processes of FRN are operated on each channel respectively. Therefore, we adopt an Inception-style synthesis method on the concatenated feature maps to composite multiple features together and reduce the number of channels and scale of concatenated feature maps simultaneously.
	
\begin{figure}[!htb]
	\centering
	\includegraphics[width=0.5\textwidth]{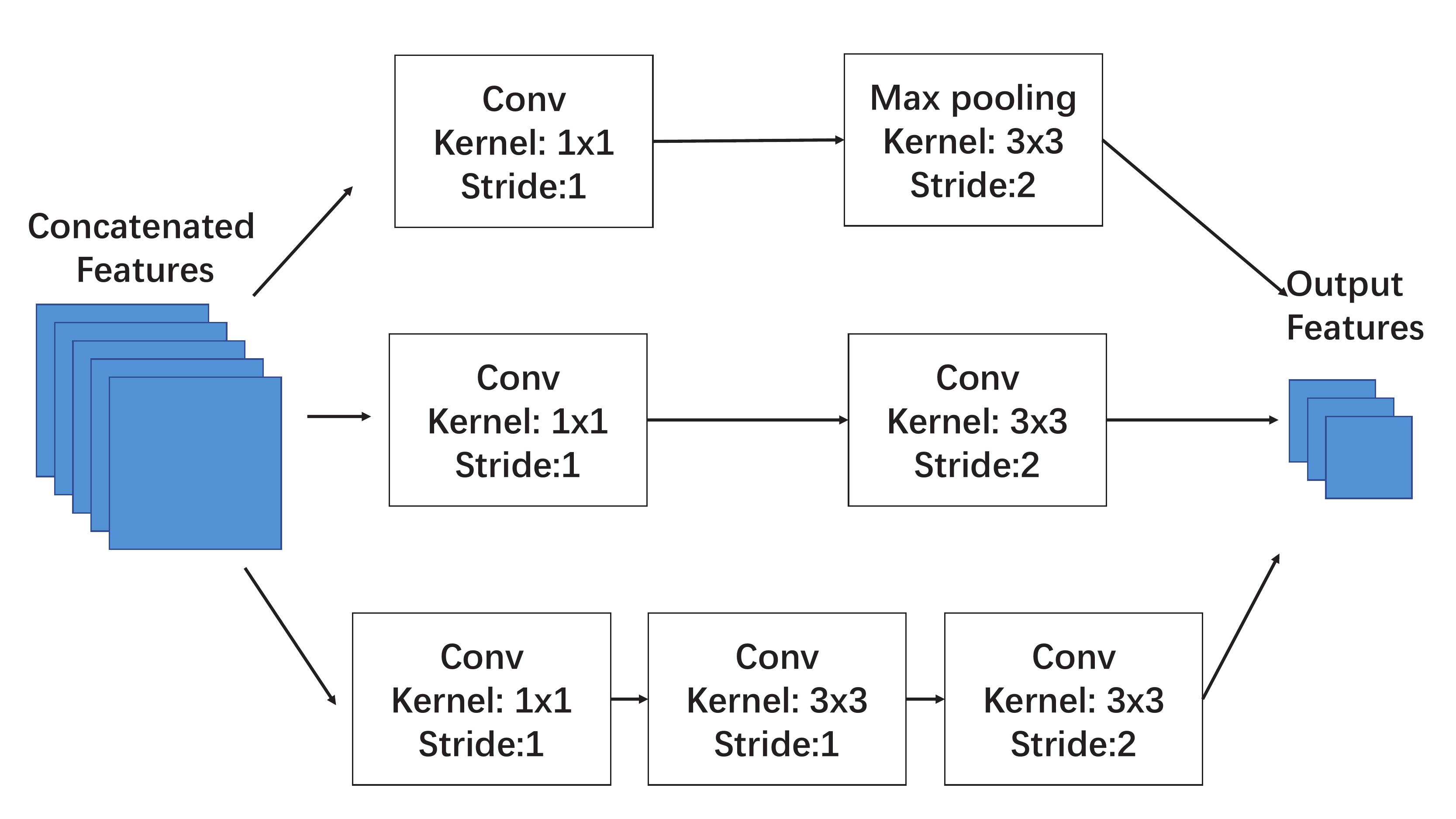}
	\caption{Inception-style synthesis: the overall structure is divided into three paths: (a) convolution (kernel: 1x1, stride: 1), maxpooling (kernel: 2x2, stride: 1); (b) convolution (kernel: 1x1, stride: 1), convolution (kernel: 3x3, stride: 2); (c) convolution (kernel: 1x1, stride:1), convolution (kernel: 3x3, stride: 1), convolution (kernel: 3x3, stride: 2). The outputs of these three paths are concatenated finally.}\label{inception-style systhesis}
\end{figure}
	
Inception module\cite{inception} has achieved great success on computer vision tasks because of its efficient and delicate design. Our synthesis method depicted in Fig.\ref{inception-style systhesis} takes advantage of Inception module. After Inception-style synthesis, the number of channels decreases from 3584 to 1024 and the scale of feature maps is reduced by 50\%.

\subsection{Cascaded Multi-Scale Architecture}
\label{sec:cascaded_multi-scale_architecture}
\begin{figure}[!htb]
	\centering
		\includegraphics[width=0.5\textwidth]{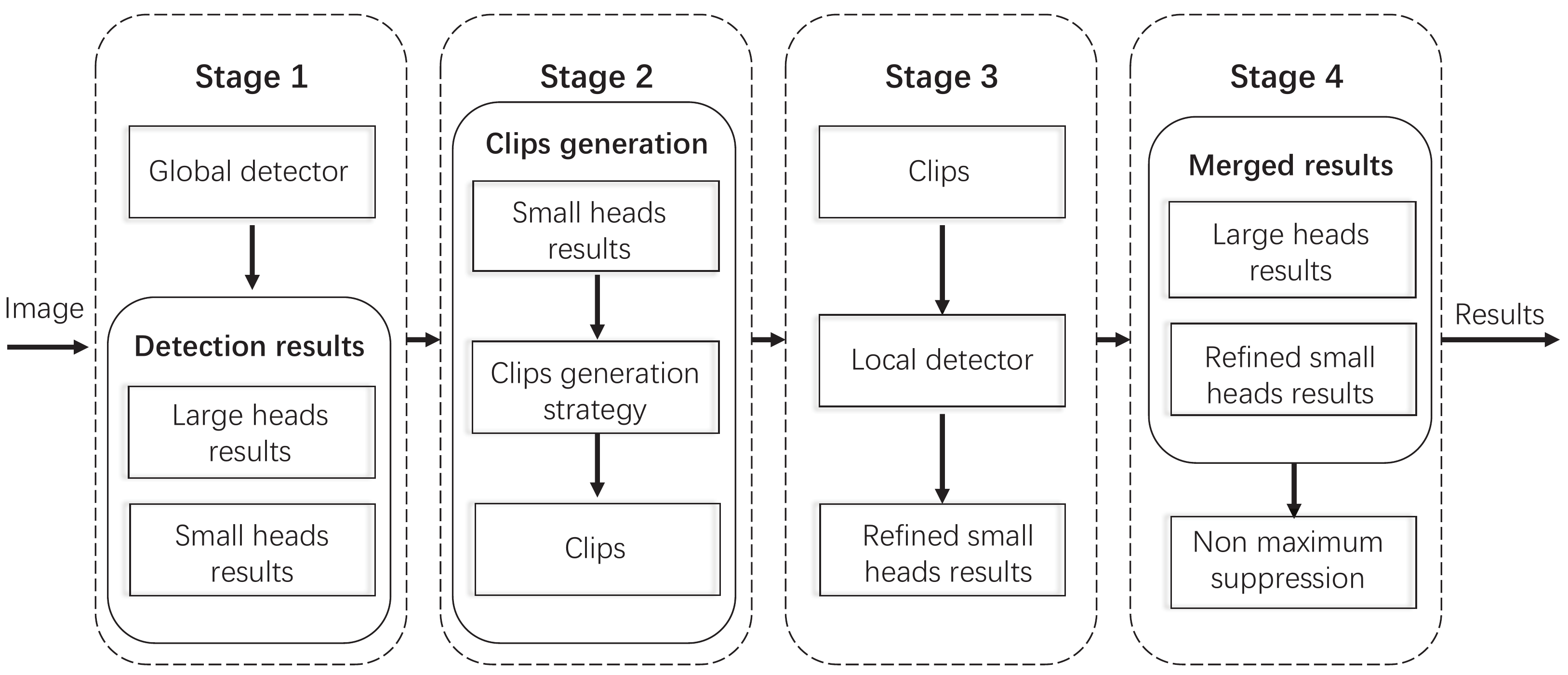}
		\caption{Procedure of Cascaded Multi-Scale Architecture: (1) global detector works on the entire image to detect large heads and obtain the rough location of small heads; (2) multiple clips which have high probability of containing small heads are generated and enlarged; (3) local detector works on the clips and results in more accurate head detection; (4) results of both detectors are merged and non maximum suppression is applied.}\label{cma}
\end{figure}

Small object detection has always been one of the most challenging problems. Previous methods\cite{scaleface, ssd, HR, perceptualgan} focus on making use of features with small receptive field or the resolution of images and feature maps to solve this problem. Inspired by the attention mechanism of humans, we propose a method named Cascaded Multi-scale Architecture for small head detection. The procedure is shown in Fig.\ref{cma}. The proposed architecture consists of two detectors named global detector and local detector, both of which are R-FCN combined with FRN.

At training stage, global detector and local detector are trained separately. The main difference in training strategies is the dataset. The global detector is trained on the original dataset while the local detector is trained on the dataset generated from the original dataset. The newly generated dataset aims to small head detection. For each image in the original dataset, we crop a $w \times w$ clip centered at each small head annotation and reserve the small head annotations whose overlaps with the clip are more than 90\%. Then, all the clips are resized to $f$ times larger, yielding the new dataset for the local detector.

At testing stage, the global detector is applied on the original image and produces the coordinates of big heads and the rough location of small heads. Then, multiple $w \times w$ clips are cropped from the input image. The clips are resized to $f$ times larger and used as the input of local detector. Local detector produces better detection of small heads. Finally, the outputs of both detectors are merged and non maximum suppression is applied on the merged results.

\begin{figure}
	\centering
	\includegraphics[width=0.3\textwidth]{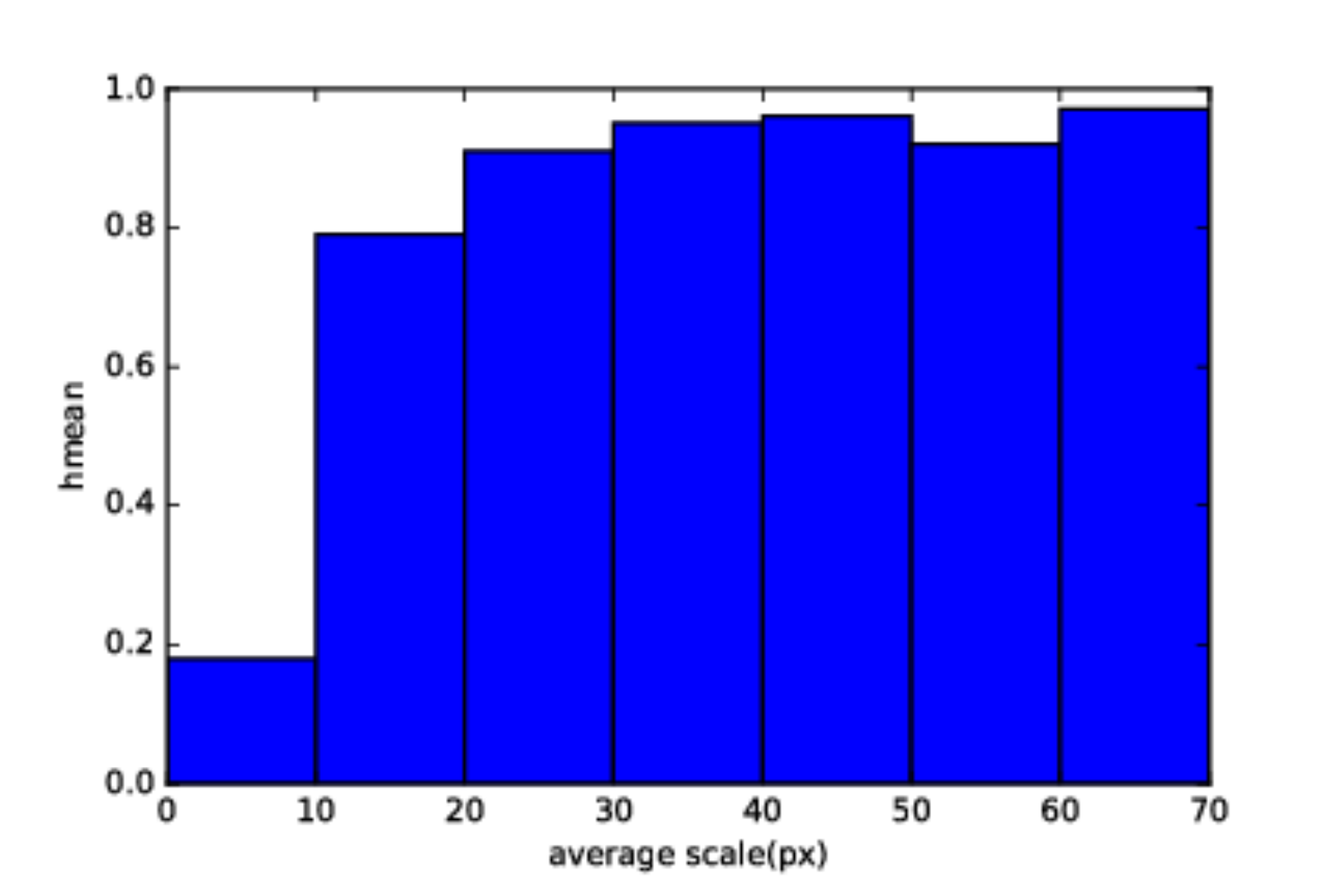}
	\caption{Detection results of different average scales ranging from $0px$ to $70px$}\label{results of different scales}
\end{figure}

The description above shows the strategy of our cascaded multi-scale architecture. However, there are also some important issues to be addressed:

\subsubsection{How to distinguish small heads?}
We define the average scale of an annotation as the average of its width and height. The performance of R-FCN with FRN w.r.t different average scales ranging from $0 px$ to $70 px$ is shown in Fig.\ref{results of different scales}. The Hmean result of heads with average scale less than $20 px$ is much lower than heads with larger scales. Therefore, we regard heads with average scale less than $20 px$ as small heads and others as large heads.

\subsubsection{How to determine the zooming factor $f$?}
The average scales of small heads range from $8 px$ to $20 px$ and the average value is around $16 px$. Our detector has a good performance on scales larger than $20 px$. Considering the computational complexity, $f$ is set to 3. Then the average scales of small heads range from $24 px$ to $60 px$ after clips are resized. Therefore, the accuracy of small heads detection can be improved.

\subsubsection{How to determine the scale of clips $w$?}
When performing cropping operation, we reserve the small head annotations whose overlaps with the clip are more than 90\%. The small head annotations whose overlaps with the clip are less than 90\% and big head annotations are abandoned. Therefore, the information contained in the overlapping area with the abandoned annotations becomes noise. To determine the scale of clips, we use a similar metric like signal-to-noise ratio in information theory. We define the area of reserved annotations as signal and the overlapping area of abandoned small heads and large heads as noise. Let $s$, $n^s$, $n^l$ denote the signal, the noise from small heads, and the noise from large heads. The scale ratio of large heads over small heads is around 4 which means $n^l$ will overwhelm $n^s$. To solve this problem, we add weights to $n^l$ and $n^s$ and set the value of weights in inverse proportion to the scale of large heads and small heads. So we set $w_l$ to 0.2 and $w_s$ to 0.8. Let C denote the number of clips and W denote the set of values of w. Then w is determined by:

\begin{equation}
w = \mathop{argmax}_{w \in W} \frac{\sum_{c=1}^{C} \frac{s_c}{w_s \cdot n_c^s + w_l \cdot n_c^l}}{C}
\end{equation}

We set $W$ = \{64, 80, 96, 112, 128, 144, 160, 176\}. The biggest signal-to-noise ratio is 3.626 when $w$ = 112. Therefore, the best choice of $w$ is 112. 

\subsubsection{How to generate clips at testing stage?}
At testing stage, we cannot crop a clip for each small heads in consideration of efficiency. Let B denote the small head detection results of global detector. For every bounding box in set B, we crop a $w \times w$ clip centered at this bounding box and delete all the bounding boxes reserved in the clip from set B. The cropping operation is not finished until set B is empty. 

\begin{table}[!htb]
	\centering
	\caption{Comparation between previous methods and our method(Multi-Scale denotes the cascaded architecture based on R-FCN + FRN)}\label{comparation between methods}
	\begin{tabular}{c|ccc|ccc}
		\hline
		Method & \multicolumn{3}{c}{PartA} & \multicolumn{3}{c}{PartB} \\
		\hline
		& P & R &  H & P & R & H  \\
		\hline
		Faster R-CNN\cite{faster-rcnn}(VGG16) & 0.86 & 0.78 & 0.82 & 0.87 & 0.81 & 0.84\\
		\hline
		YOLOv2\cite{yolo9000} & 0.91 & 0.61 & 0.73 & 0.74 & 0.67 & 0.70\\
		\hline
		SSD\cite{ssd} & 0.84 & 0.68 & 0.76 & 0.80 & 0.66 & 0.72\\
		\hline
		R-FCN\cite{rfcn} (ResNet-50) & 0.87 & 0.78 & 0.82 & 0.90 & 0.82 & 0.86\\
		\hline
		\tabincell{c}{R-FCN + FRN (ResNet-50) \\ (proposed)} & 0.89 & 0.83 & 0.86 & 0.92 & 0.84 & 0.88 \\
		\hline
		Multi-Scale (proposed)& \textbf{0.92} & \textbf{0.90} & \textbf{0.91} & \textbf{0.92} & \textbf{0.87} & \textbf{0.90} \\
		\hline
	\end{tabular}	
\end{table}

\section{Experiments}
\label{sec:experiments}
	
\subsection{Datasets}
\begin{figure}[!hbt]
	\centering
	\subfigure[]{\includegraphics[height=0.1\textheight]{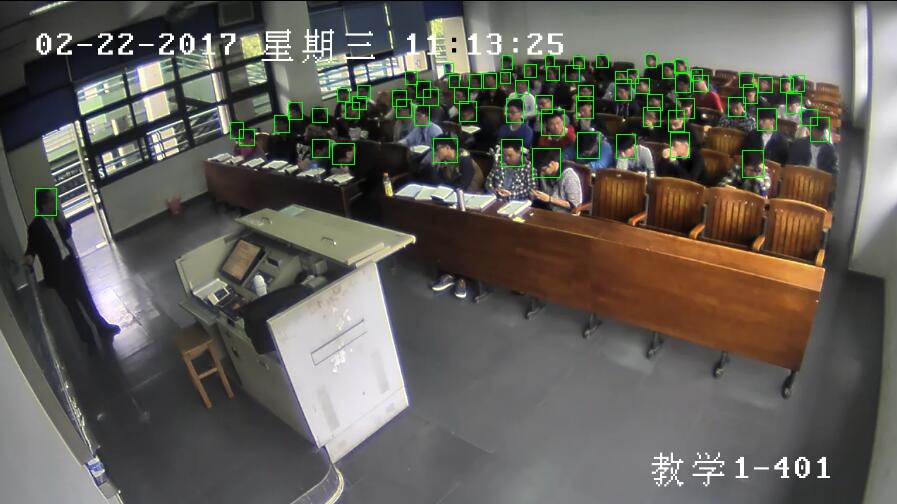}}
	\subfigure[]{\includegraphics[height=0.1\textheight]{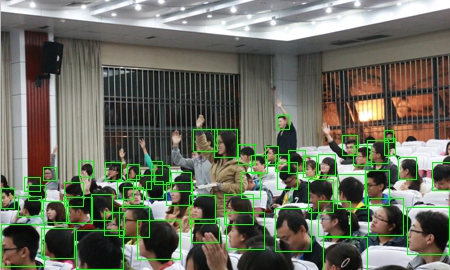}}
	\caption{(a) An example image and annotations of PartA in SCUT-HEAD. (b) An example image and annotations of PartB in SCUT-HEAD.}
	\label{representative images and annotations}
	\end{figure}

We have collected and labeled a large-scale head detection dataset named SCUT-HEAD\footnote[1]{The SCUT-HEAD dataset can be downloaded from https://github.com /HCIILAB/SCUT-HEAD-Dataset-Release.}. The proposed dataset consists of two parts. PartA includes 2000 images sampled from monitor videos of classrooms in a university with 67321 heads annotated. As classrooms of one university usually look similar and the poses of people vary less, we carefully choose representative images to gain variance and reduce similarity.  PartB includes 2405 images crawled from the Internet with 43930 heads annotated. We have labeled every visible head with xmin, ymin, xmax and ymax coordinates and ensured that annotations cover the entire head including the blocked parts but without extra background. Both PartA and PartB are divided into training and testing parts. 1500 images of PartA are for training and 500 for testing. 1905 images of PartB are for training and 500 for testing. Our dataset follows the standard of Pascal VOC. Two representative images and annotations are shown in Fig.\ref{representative images and annotations}. 
	
\subsection{Implementation details}
Global detector and local detector are trained using stochastic gradient descent (SGD). Momentum and weight decay are set to be 0.9 and 0.0005 respectively. Widths of images are resized to 1024 while keeping their aspect ratios. Learning rate is set to 0.001 during 0 $\sim$ 10k iterations, 0.0001 during 10k $\sim$ 20k iterations and 0.00001 during 20k $\sim$ 30k iterations. As for anchors setting strategy, we generate anchors using Kmeans with modified distance metrics\cite{yolo9000}. Online hard example mining (OHEM)\cite{ohem} is also applied for more effective training.

\subsection{Results}
\label{results}
We compare our method with other object detection methods. Results in Table \ref{comparation between methods} imply that our method has a great improvement compared to other methods especially after applying the cascaded multi-scale architecture. We have also compared the performance of R-FCN using different techniques of FRN in Table \ref{comparation in FRN}, which proves that our final design of FRN reaches the best result. Note that the method without Inception-style synthesis replaces it by convolution layer (kernel size: 1x1, stride: 1) followed by max pooling (kernel size: 2, stride: 2) and method without feature decomposition upsampling replaces it by bilinear interpolation.
	
\begin{table}[!htb]
	\centering
	\caption{Comparation between using different techniques of FRN (Based on PartA of SCUT-HEAD)}\label{comparation in FRN}
	\begin{tabular}{c|ccc}
		\hline
		Technique & \multicolumn{3}{c}{Usage} \\
		\hline
		Channel weighting  & $\surd$ & $\surd$ & $\surd$ \\
		\hline
		Inception-style synthesis  & & $\surd$ & $\surd$ \\
		\hline
		Feature decomposition upsampling  & & & $\surd$ \\
		\hline
		Result (Hmeans)  & 0.8412 & 0.8497 & \textbf{0.8591}\\
		\hline
	\end{tabular}	
\end{table}
	
\subsection{Channel weighting}
\label{channel weighting}
\begin{figure}
	\centering
	\includegraphics[width=0.5\textwidth]{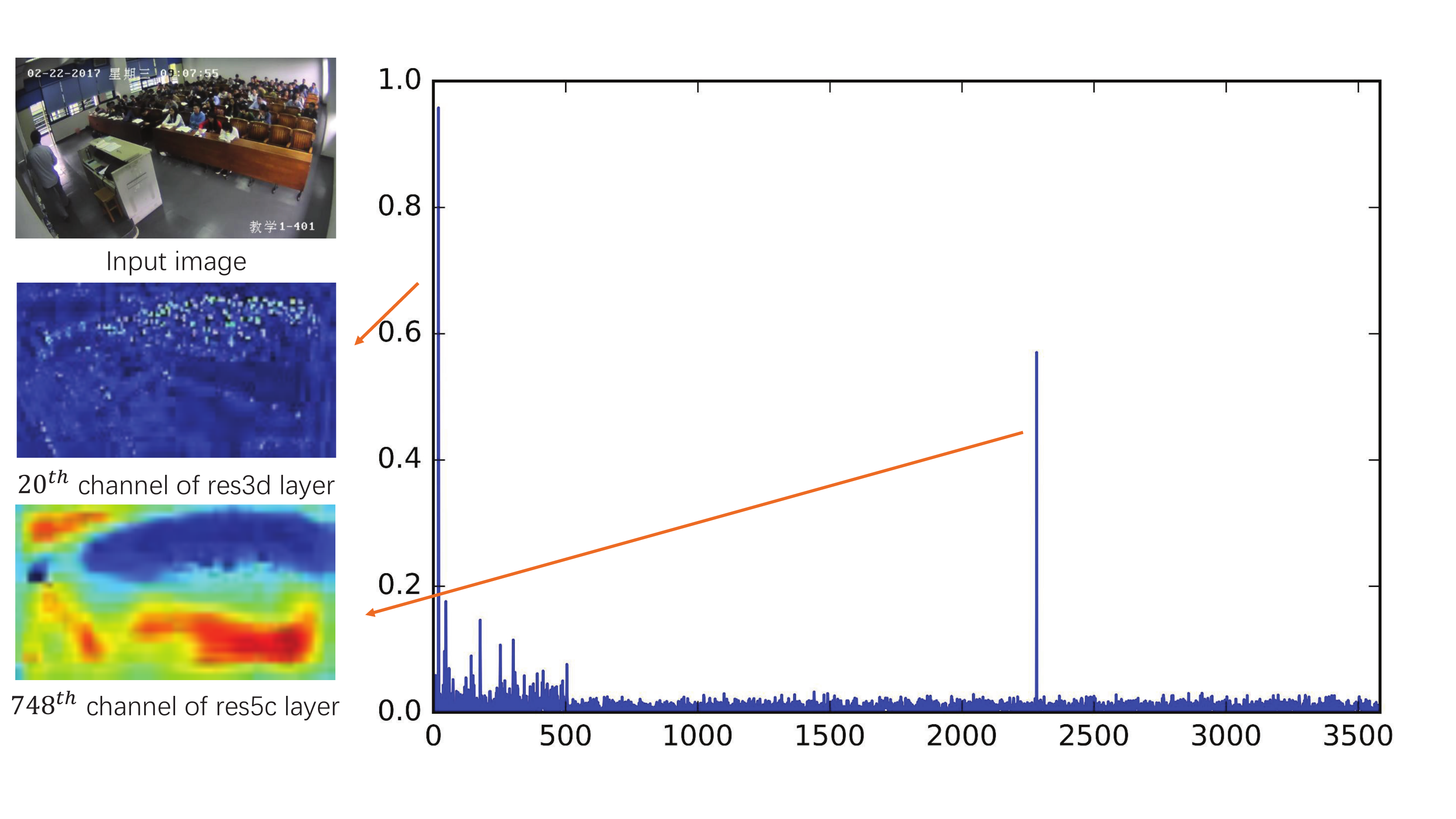}
	\caption{Weights for channels of res3, res4 and res5 ([0, 511] for res3, [513, 1535] for res4 and [1536, 3583] for res5) and visualization of two features with biggest weights.}\label{wc}
\end{figure}

We plot the weights of channel weighting layers in Fig.\ref{wc} to indicate the importance of different features. The two channels with the biggest weights are the \textsl{20th} channel of res3 and the \textsl{748th} channel of res5. From the visualization of these two channels and the input image, we can have a better understanding of the CNN and effectiveness of channel weighting. The \textsl{20th} channel of res3 is better at indicating the location of heads. The location of heads can be easily estimated by light blue points. The \textsl{748th} channel of res5 is more suitable for the classification of heads and background. Heads are indicated by the blue areas and the remaining areas indicate the background. It is implied that the feature maps at low level localize objects more precisely while the feature maps at high level do better in classification.

\begin{figure}
	\centering
	\includegraphics[width=0.08\textwidth]{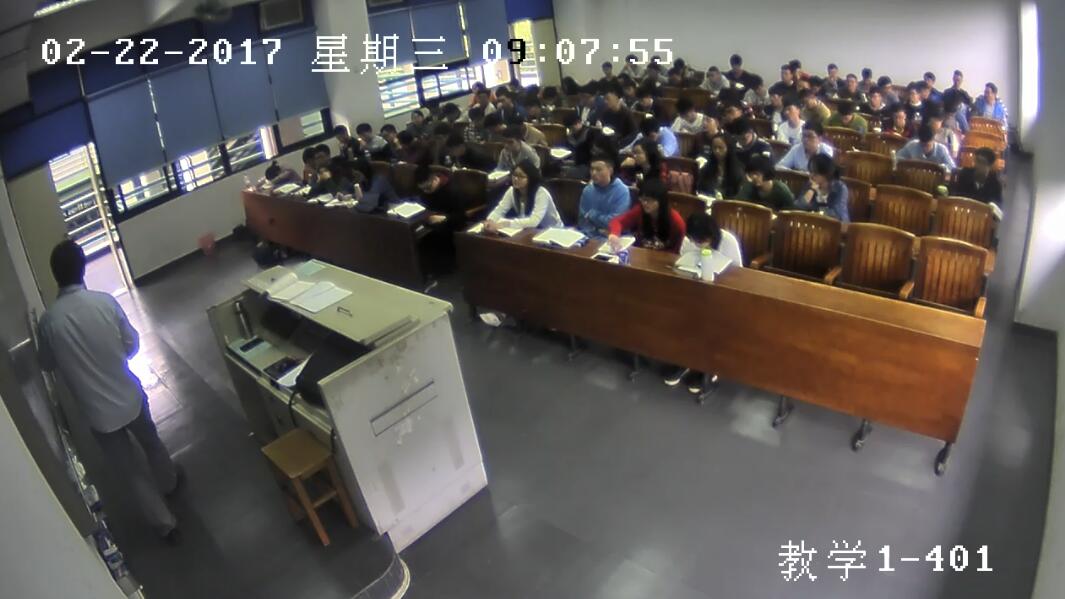}
	\includegraphics[width=0.08\textwidth]{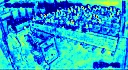}
	\includegraphics[width=0.08\textwidth]{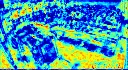}
	\includegraphics[width=0.08\textwidth]{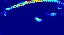}
	\includegraphics[width=0.08\textwidth]{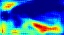}
	\caption{Visualization of the input image and some other feature maps from res3 and res5 (the left two are from res3 and the right two are from res5).}\label{other feature maps}
\end{figure}

Some other randomly selected feature maps are visualized in Fig.\ref{other feature maps}. Compared to the two feature maps shown in Fig.\ref{wc}, the randomly selected feature maps don't have implicit relationship with the goal of head detection.
	
%From the visualization of these two features, we can acquire a general distribution of heads by light blue points in the \textsl{20th} feature of res3d and the blue areas in the \textsl{748th} feature of res5c correspond to the area containing heads and the other area corresponds to the background.
	
\begin{figure}
	\centering
	\includegraphics[width=0.5\textwidth]{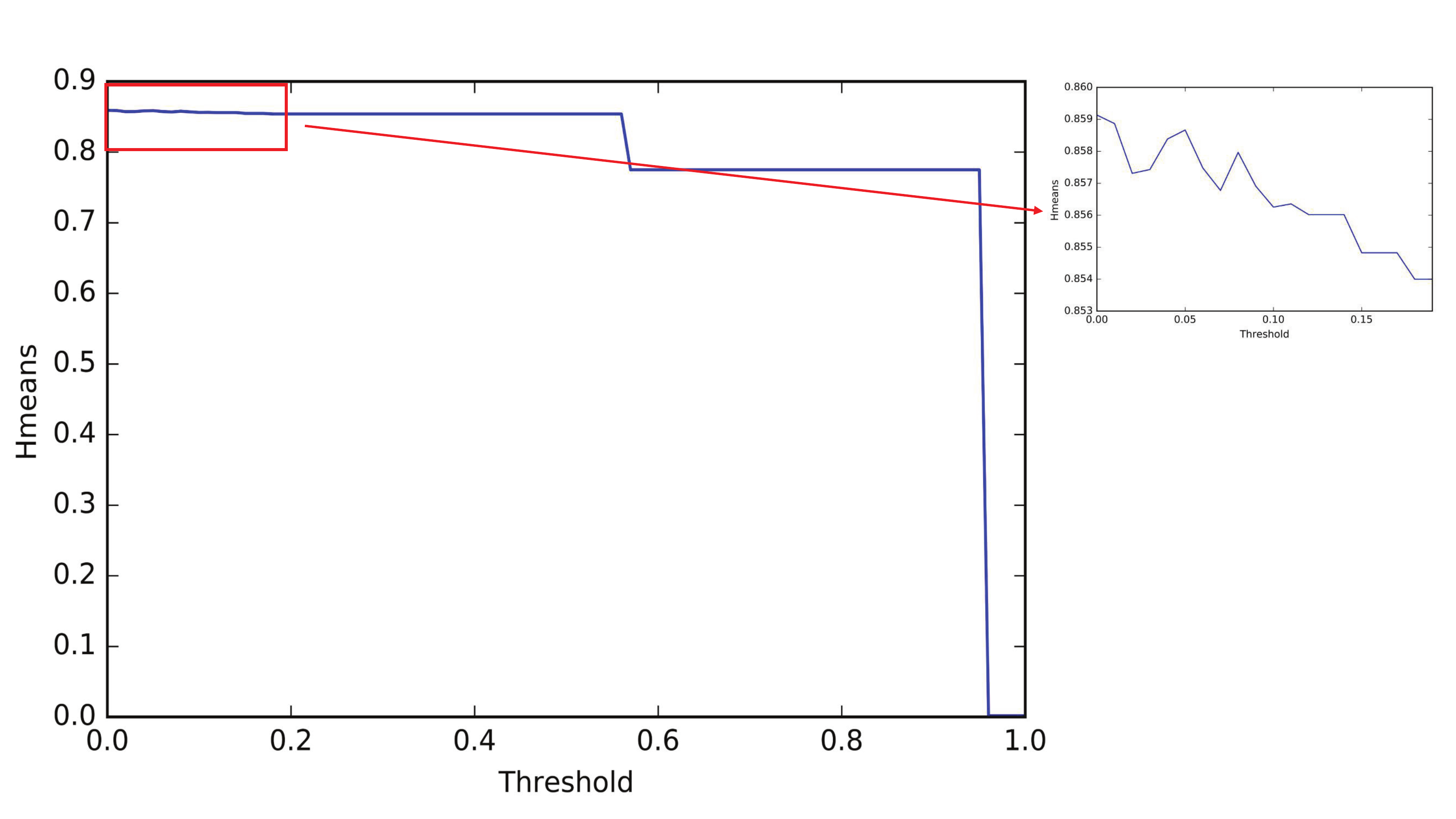}
	\caption{Detection result using features with weight bigger than a threshold (weights of unused features are set to zero). It implies that the features with big weights contribute much to the final result while the features with small weights do not really make sense.}\label{comparation in weight}
\end{figure}
	
The results using features with weights bigger than a threshold are shown in Fig.\ref{comparation in weight}.  When threshold ranges from 0.00 to 0.20, some features with small weights are abandoned. However, it has a really small influence on the performance of the whole detection framework. Although all the features with weights smaller than 0.20 are abandoned, the performance only decreases by 0.05 which is negligible. The whole curve only has two sharp drops. The first drop happens after the \textsl{748th} channel of res5 is abandoned. The second drop happens after the  \textsl{20th} channel of res3 is abandoned. The performance drops to nearly zero. It is implied that our channel weighting layers use features effectively and alternatively towards the goal of head detection.

From the above analysis, we can conclude that channel weighting performs feature selection very well. The most useful features for the specific goal of head detection are selected and made use of.

\subsection{Small head detection}
We improve small head detection through two ways. The first is FRN which combines multiple features at different levels together. The feature maps at low levels have smaller receptive field and more detailed information which are beneficial to small head detection. The second is the cascaded multi-scale architecture. It is designed for small head detection specifically through the combination of global and local information. The results of small heads detection are shown in Table \ref{small head detection performance}.
\begin{table}
	\caption{Small head detection performances}\label{small head detection performance}
	\centering
	\begin{tabular}{c|ccc|ccc}
		\hline
		Average scale & \multicolumn{3}{c}{0 $\sim$ 10 \textsl{px}} \vline & \multicolumn{3}{c}{10 $\sim$ 20 \textsl{px}} \\
		\hline
		Method & P & R & H & P & R & H  \\
		\hline
		R-FCN & 0.12 & 0.10 & 0.11 & 0.53 & 0.77 & 0.63\\
		\hline
		R-FCN + FRN & 0.17 & 0.19 & 0.18 & 0.83 & 0.76 & 0.80\\
		\hline
		Multi-scale & \textbf{0.63} & \textbf{0.57} & \textbf{0.60} & \textbf{0.93} & \textbf{0.84} & \textbf{0.88} \\
		\hline
	\end{tabular}
\end{table}
	
%	\left%
%	\includegraphics[width=5cm]{weights.pdf}
%	\caption{Weights for channels of feature maps produced by res3d, res4f, res5c layers ([0, 511] for res3d layer, [513: 1535] for res4f layer and [1536:3583] for res5c layer)}\label{inception-style systhesis}
%\begin{table}[!htb]
%	\centering
%	\caption{Detection result using features with weight bigger than a threshold}\label{comparation in weight}
%	\begin{tabular}{c|ccccc}
%		\hline
%		Threshold & 0.0 & 0.5 & 0.3 & 0.2 & 0.1 \\
%		Precision & & & & & \\
%		\hline 
%		Recall & & & & & \\
%		\hline
%		Hmeans & & & & & \\
%		\hline
%	\end{tabular}	
%\end{table}

\subsection{Other datasets}
We also compare our method on Brainwash dataset\cite{brainwash} in Table \ref{comparation on brainwash}. Brainwash dataset contains 91146 heads annotated in 11917 images. Our method also achieves state-of-the-art performance on this dataset compared with several baselines including context-aware CNNs local model (Con-local)\cite{context}, end-to-end people detection with Hungarian loss (ETE-hung)\cite{brainwash}, localized fusion method (f-localized)\cite{localized} and R-FCN\cite{rfcn}.

\begin{table}[!htb]
	\centering
\caption{Comparation on Brainwash dataset}\label{comparation on brainwash}
\begin{tabular}{cccccc}
	\hline
	Method&Con-local&ETE-hung&R-FCN&f-localized&our method \\
	\hline
	AP & 45.4 & 78.4 & 84.8 & 85.3 &\textbf{88.1}\\
	\hline
\end{tabular}	
 
\end{table}

\section{Conclusion}
In this paper, we have proposed a novel method for head detection using Feature Refine Net (FRN) and cascaded multi-scale architecture. FRN combines multi-scale features and takes advantage of the most useful features. The cascaded multi-scale architecture focuses on small heads detection. Owing to these techniques, our method achieves a great performance on indoor-scene head detection. Furthermore, we built a dataset named SCUT-HEAD which is for indoor-scene head detection. 

\section{Acknowledgement}
This research is supported in part by the National Key R\&D Program of China (No. 2016YFB1001405), NSFC (Grant\\No.: 61472144, 61673182, 61771199), GD-NSF (no. 2017A0\\30312006), GDSTP (Grant No.: 2014A010103012, 2015B01\\0101004, 2015B010130003), GZSTP (no. 201607010227).


\begin{thebibliography}{10}
\providecommand{\url}[1]{#1}
\csname url@samestyle\endcsname
\providecommand{\newblock}{\relax}
\providecommand{\bibinfo}[2]{#2}
\providecommand{\BIBentrySTDinterwordspacing}{\spaceskip=0pt\relax}
\providecommand{\BIBentryALTinterwordstretchfactor}{4}
\providecommand{\BIBentryALTinterwordspacing}{\spaceskip=\fontdimen2\font plus
\BIBentryALTinterwordstretchfactor\fontdimen3\font minus
  \fontdimen4\font\relax}
\providecommand{\BIBforeignlanguage}[2]{{%
\expandafter\ifx\csname l@#1\endcsname\relax
\typeout{** WARNING: IEEEtran.bst: No hyphenation pattern has been}%
\typeout{** loaded for the language `#1'. Using the pattern for}%
\typeout{** the default language instead.}%
\else
\language=\csname l@#1\endcsname
\fi
#2}}
\providecommand{\BIBdecl}{\relax}
\BIBdecl

\bibitem{hypercolumn}
B.~Hariharan, P.~Arbel{\'a}ez, R.~Girshick, and J.~Malik, ``Hypercolumns for
  object segmentation and fine-grained localization,'' in \emph{Proceedings of
  the IEEE Conference on Computer Vision and Pattern Recognition}, 2015, pp.
  447--456.

\bibitem{ssd}
W.~Liu, D.~Anguelov, D.~Erhan, C.~Szegedy, S.~Reed, C.-Y. Fu, and A.~C. Berg,
  ``Ssd: Single shot multibox detector,'' in \emph{European conference on
  computer vision}.\hskip 1em plus 0.5em minus 0.4em\relax Springer, 2016, pp.
  21--37.

\bibitem{fpn}
T.-Y. Lin, P.~Doll{\'a}r, R.~Girshick, K.~He, B.~Hariharan, and S.~Belongie,
  ``Feature pyramid networks for object detection,'' \emph{arXiv preprint
  arXiv:1612.03144}, 2016.

\bibitem{hypernet}
T.~Kong, A.~Yao, Y.~Chen, and F.~Sun, ``Hypernet: Towards accurate region
  proposal generation and joint object detection,'' in \emph{Proceedings of the
  IEEE Conference on Computer Vision and Pattern Recognition}, 2016, pp.
  845--853.

\bibitem{parsenet}
W.~Liu, A.~Rabinovich, and A.~C. Berg, ``Parsenet: Looking wider to see
  better,'' \emph{arXiv preprint arXiv:1506.04579}, 2015.

\bibitem{HR}
P.~Hu and D.~Ramanan, ``Finding tiny faces,'' in \emph{2017 IEEE Conference on
  Computer Vision and Pattern Recognition (CVPR)}.\hskip 1em plus 0.5em minus
  0.4em\relax IEEE, 2017, pp. 1522--1530.

\bibitem{faster-rcnn}
S.~Ren, K.~He, R.~Girshick, and J.~Sun, ``Faster r-cnn: Towards real-time
  object detection with region proposal networks,'' in \emph{Advances in neural
  information processing systems}, 2015, pp. 91--99.

\bibitem{rfcn}
J.~Dai, Y.~Li, K.~He, and J.~Sun, ``R-fcn: Object detection via region-based
  fully convolutional networks,'' in \emph{Advances in neural information
  processing systems}, 2016, pp. 379--387.

\bibitem{yolo9000}
J.~Redmon and A.~Farhadi, ``Yolo9000: better, faster, stronger,'' \emph{arXiv
  preprint arXiv:1612.08242}, 2016.

\bibitem{imagenet}
J.~Deng, W.~Dong, R.~Socher, L.-J. Li, K.~Li, and L.~Fei-Fei, ``Imagenet: A
  large-scale hierarchical image database,'' in \emph{Computer Vision and
  Pattern Recognition, 2009. CVPR 2009. IEEE Conference on}.\hskip 1em plus
  0.5em minus 0.4em\relax IEEE, 2009, pp. 248--255.

\bibitem{vgg}
K.~Simonyan and A.~Zisserman, ``Very deep convolutional networks for
  large-scale image recognition,'' \emph{arXiv preprint arXiv:1409.1556}, 2014.

\bibitem{resnet}
K.~He, X.~Zhang, S.~Ren, and J.~Sun, ``Deep residual learning for image
  recognition,'' in \emph{Proceedings of the IEEE conference on computer vision
  and pattern recognition}, 2016, pp. 770--778.

\bibitem{inception}
C.~Szegedy, W.~Liu, Y.~Jia, P.~Sermanet, S.~Reed, D.~Anguelov, D.~Erhan,
  V.~Vanhoucke, and A.~Rabinovich, ``Going deeper with convolutions,'' in
  \emph{Proceedings of the IEEE conference on computer vision and pattern
  recognition}, 2015, pp. 1--9.

\bibitem{scaleface}
S.~Yang, Y.~Xiong, C.~C. Loy, and X.~Tang, ``Face detection through
  scale-friendly deep convolutional networks,'' \emph{arXiv preprint
  arXiv:1706.02863}, 2017.

\bibitem{perceptualgan}
J.~Li, X.~Liang, Y.~Wei, T.~Xu, J.~Feng, and S.~Yan, ``Perceptual generative
  adversarial networks for small object detection,'' in \emph{IEEE CVPR}, 2017.

\bibitem{ohem}
A.~Shrivastava, A.~Gupta, and R.~Girshick, ``Training region-based object
  detectors with online hard example mining,'' in \emph{Proceedings of the IEEE
  Conference on Computer Vision and Pattern Recognition}, 2016, pp. 761--769.

\bibitem{brainwash}
R.~Stewart, M.~Andriluka, and A.~Y. Ng, ``End-to-end people detection in
  crowded scenes,'' in \emph{Proceedings of the IEEE Conference on Computer
  Vision and Pattern Recognition}, 2016, pp. 2325--2333.

\bibitem{context}
T.-H. Vu, A.~Osokin, and I.~Laptev, ``Context-aware cnns for person head
  detection,'' in \emph{Proceedings of the IEEE International Conference on
  Computer Vision}, 2015, pp. 2893--2901.

\bibitem{localized}
Y.~Li, Y.~Dou, X.~Liu, and T.~Li, ``Localized region context and object feature
  fusion for people head detection,'' in \emph{Image Processing (ICIP), 2016
  IEEE International Conference on}.\hskip 1em plus 0.5em minus 0.4em\relax
  IEEE, 2016, pp. 594--598.

\end{thebibliography}
\end{document}